\pdfoutput=1
\documentclass[11pt]{article}

\usepackage{acl}

\usepackage{times}
\usepackage{latexsym}

\usepackage{dsfont}

\usepackage{xfrac}
\usepackage{cancel}

\usepackage{emoji}

\usepackage{mathtools}

\allowdisplaybreaks

\usepackage[T1]{fontenc}
\usepackage{microtype}
\everypar{\looseness=-1}
\usepackage{inconsolata}

\usepackage{booktabs}
\usepackage{amsmath,amssymb,amsthm,amsfonts}
\usepackage{pifont}
\usepackage{xcolor}
\usepackage{listings}
\usepackage[scaled]{beramono}

\usepackage{makecell}
\usepackage{graphicx}
\usepackage[export]{adjustbox}

\usepackage{xspace}

\usepackage{color}
\usepackage{colortbl}
\usepackage{multirow}
\usepackage{bbm}
\usepackage{enumitem}
\usepackage{verbatimbox}
\usepackage{cleveref}
\crefname{section}{§}{§§}
\Crefname{section}{§}{§§}

\usepackage[belowskip=-10pt,aboveskip=10pt]{caption}

\usepackage{relsize}
\newcommand{\rug}{\emoji[openmoji]{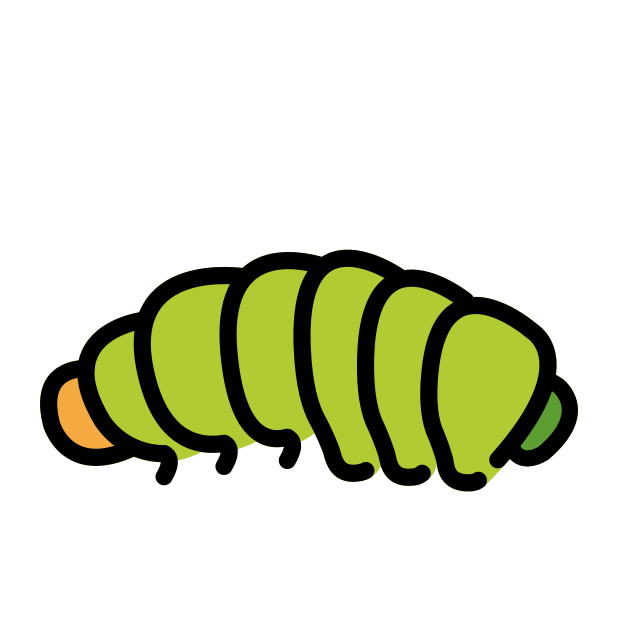}}
\newcommand{\dfki}{\emoji[openmoji]{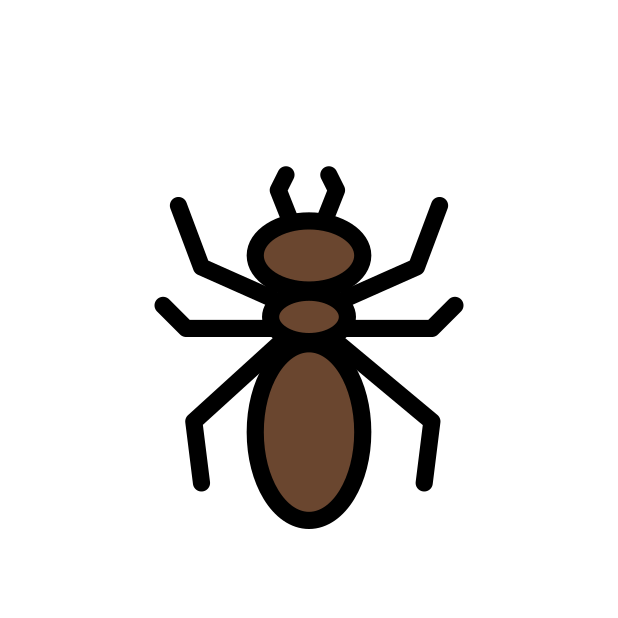}}
\newcommand{\uva}{\emoji[openmoji]{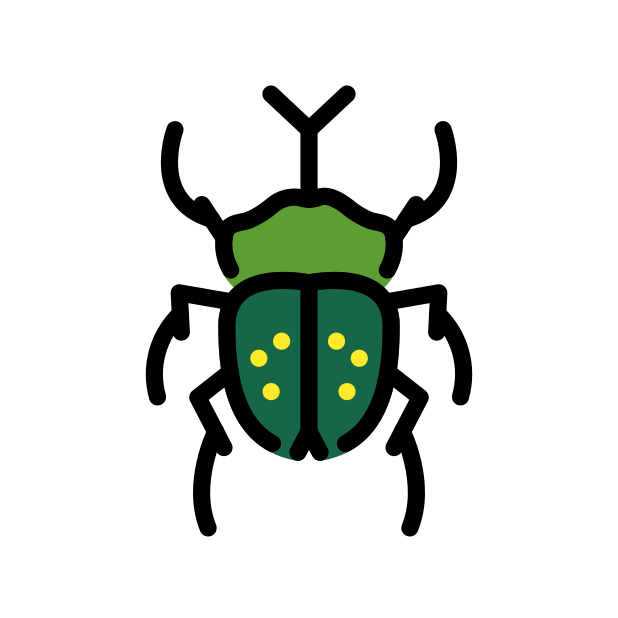}}
\newcommand{\hf}{\emoji[openmoji]{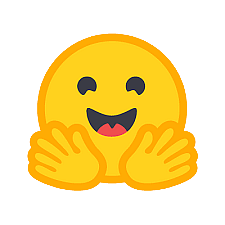}}
\newcommand{\male}{\emoji[openmoji]{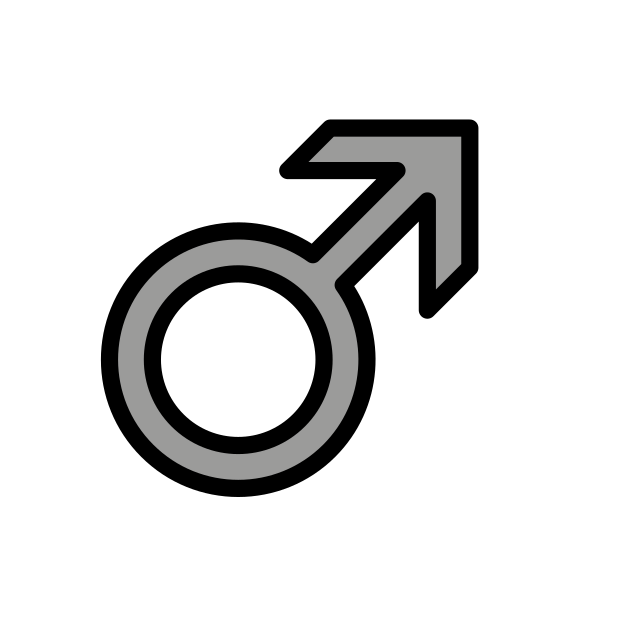}}
\newcommand{\female}{\emoji[openmoji]{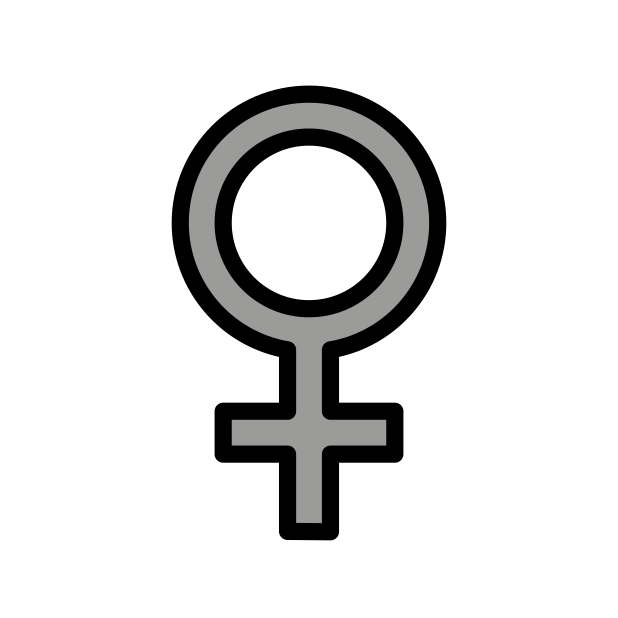}}
\newcommand{\yes}{{\color{teal}\ding{51}}}
\newcommand{\no}{{\color{purple}\ding{55}}}

\definecolor{inseqred}{RGB}{188, 62, 68}
\definecolor{codegreen}{rgb}{0,0.6,0}
\definecolor{codegray}{rgb}{0.5,0.5,0.5}

\definecolor{backcolour}{RGB}{245,248,250}
\definecolor{emph}{RGB}{166,88,53}
\definecolor{nightblue}{RGB}{9,49,105}
\definecolor{keywords}{RGB}{207,33,46}
\definecolor{lightpurple}{RGB}{130,81,223}
\definecolor{lightredclear}{RGB}{234,153,153}
\definecolor{lightreddim}{RGB}{224,102,102}
\definecolor{lightblueclear}{RGB}{159,197,232}
\definecolor{lightbluedim}{RGB}{111,168,220}

\newcommand{\sig}{$^*$}
\newcommand{\notsig}{$^{\phantom{*}}$}

\lstdefinestyle{fullpagecode}{
    backgroundcolor=\color{backcolour},   
    commentstyle=\color{codegreen},
    keywordstyle=\color{keywords},
    stringstyle=\color{nightblue},
    basicstyle=\ttfamily\footnotesize,
    breakatwhitespace=false,         
    breaklines=true,                 
    captionpos=b,                    
    keepspaces=true,             
    showspaces=false,                
    showstringspaces=false,
    showtabs=false,                  
    tabsize=2,
    frame=shadowbox,
    emph={inseq, True, AggregatorPipeline, ContiguousSpanAggregator, SequenceAttributionAggregator, PairAggregator, AutoModelForCausalLM, AutoTokenizer, datasets, transformers, SubwordAggregator, TunedLens, tuned_lens},
    emphstyle={\color{emph}},
    emph={[2]load_model,attribute,show, aggregate, load_dataset, encode, from_pretrained, load, confidence_from_prediction_depth},
    emphstyle={[2]\color{lightpurple}},
    linewidth=\textwidth,
    extendedchars=true,
    literate={ş}{{\c{s}}}1
}

\lstdefinestyle{singlecolumncode}{
    backgroundcolor=\color{backcolour},   
    commentstyle=\color{codegreen},
    keywordstyle=\color{keywords},
    stringstyle=\color{nightblue},
    basicstyle=\ttfamily\footnotesize,
    breakatwhitespace=false,         
    breaklines=true,                 
    captionpos=b,                    
    keepspaces=true,             
    showspaces=false,                
    showstringspaces=false,
    showtabs=false,                  
    tabsize=2,
    frame=shadowbox,
    emph={inseq, True, AggregatorPipeline, ContiguousSpanAggregator, SequenceAttributionAggregator, PairAggregator, SubwordAggregator, tuned_lens, TunedLens},
    emphstyle={\color{emph}},
    emph={[2]load_model,attribute,show, aggregate, load},
    emphstyle={[2]\color{lightpurple}},
    linewidth=7.5cm
}

\newcommand{\emo}[1]{\raise1.0ex\hbox{\normalfont#1}}

\renewcommand{\digamma}{\psi}

\usepackage{array}
\newcolumntype{L}[1]{>{\raggedright\let\newline\\\arraybackslash\hspace{0pt}}m{#1}}
\newcolumntype{C}[1]{>{\centering\let\newline\\\arraybackslash\hspace{0pt}}m{#1}}
\newcolumntype{R}[1]{>{\raggedleft\let\newline\\\arraybackslash\hspace{0pt}}m{#1}}

\title{
    \begin{minipage}[c]{\linewidth}
    \centering
        \begin{minipage}{1cm}\vspace{-.1cm}    \centering\scalebox{1}[1]{\includegraphics[width=0.8cm]{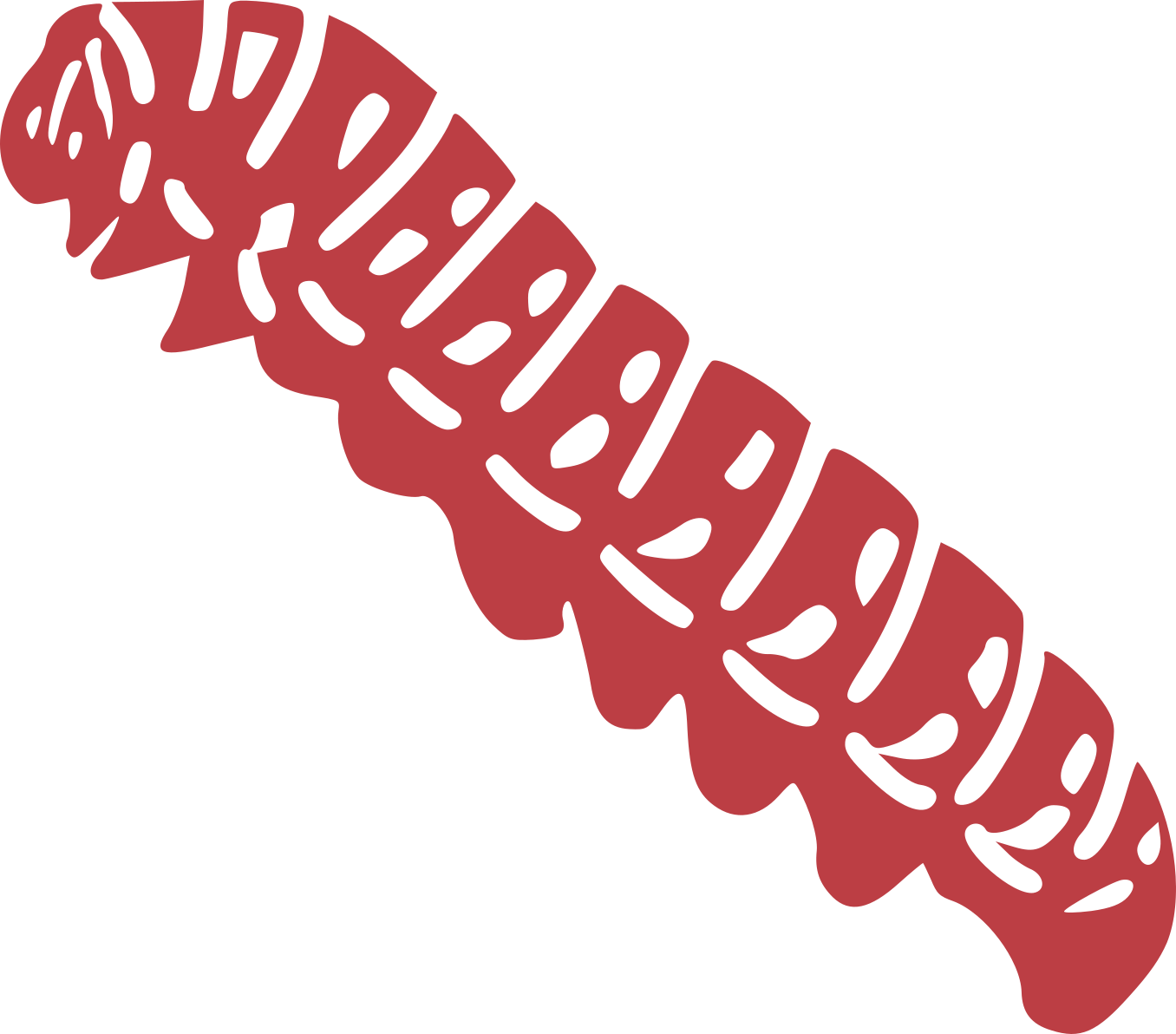}}\end{minipage}
        \textcolor{inseqred}{Inseq}: An Interpretability Toolkit for Sequence Generation Models
        \end{minipage}
}

\author{
  Gabriele Sarti~\emo{\rug}
  ~\;~
  Nils Feldhus\emo{\dfki}
  ~\;~
  Ludwig Sickert~\emo{\rug}\\
  \textbf{Oskar van der Wal}\emo{\uva}
  ~\;~
  \textbf{Malvina Nissim}~\emo{\rug}
  ~\;~
  \textbf{Arianna Bisazza}~\emo{\rug}\vspace{0.2cm}\\
  \begin{tabular}{c c}
    \emo{\rug}~University of Groningen
    \emo{\uva}University of Amsterdam
  \end{tabular}\\
  \begin{tabular}{c}
  \emo{\dfki}German Research Center for Artificial Intelligence (DFKI), Berlin
  \end{tabular}\vspace{0.2cm}\\
  \small
  \texttt{\href{mailto:g.sarti@rug.nl}{g.sarti@rug.nl}}%~\;~  \texttt{\href{mailto:nils.feldhus@dfki.de}{nils.feldhus@dfki.de}}~\;~
}

\begin{document}
\maketitle
\begin{abstract}
Past work in natural language processing interpretability focused mainly on popular classification tasks while largely overlooking generation settings, partly due to a lack of dedicated tools. In this work, we introduce Inseq\footnote{Library: \url{https://github.com/inseq-team/inseq}\\Documentation: \url{https://inseq.readthedocs.io}\\This paper describes the Inseq \href{https://github.com/inseq-team/inseq/releases/tag/v0.4.0}{\texttt{v0.4.0}} release on PyPI.}, a Python library to democratize access to interpretability analyses of sequence generation models. Inseq enables intuitive and optimized extraction of models' internal information and feature importance scores for popular decoder-only and encoder-decoder Transformers architectures. We showcase its potential by adopting it to highlight gender biases in machine translation models and locate factual knowledge inside GPT-2. Thanks to its extensible interface supporting cutting-edge techniques such as contrastive feature attribution, Inseq can drive future advances in explainable natural language generation, centralizing good practices and enabling fair and reproducible model evaluations.
\end{abstract}

\section{Introduction}
\label{sec:intro}

Recent years saw an increase in studies and tools aimed at improving our behavioral or mechanistic understanding of neural language models~\citep{belinkov-glass-2019-analysis}.
In particular, \textit{feature attribution} methods became widely adopted to quantify the importance of input tokens in relation to models' inner processing and final predictions~\citep{madsen-etal-2022-posthoc}.
Many studies applied such techniques to modern deep learning architectures, including Transformers~\citep{vaswani-etal-2017-attention}, leveraging gradients~\citep{baherens-etal-2010-explain,sundararajan-etal-2017-ig}, attention patterns~\citep{xu-etal-2015-show,clark-etal-2019-bert} and input perturbations~\citep{zeiler-fergus-2014-visualizing,feng-etal-2018-pathologies} to quantify input importance, often leading to controversial outcomes in terms of faithfulness, plausibility and overall usefulness of such explanations~\citep{adebayo-etal-2018-sanity,jain-wallace-2019-attention,jacovi-goldberg-2020-towards,zafar-etal-2021-lack}.
\begin{figure}
    \includegraphics[width=\linewidth, angle=0]{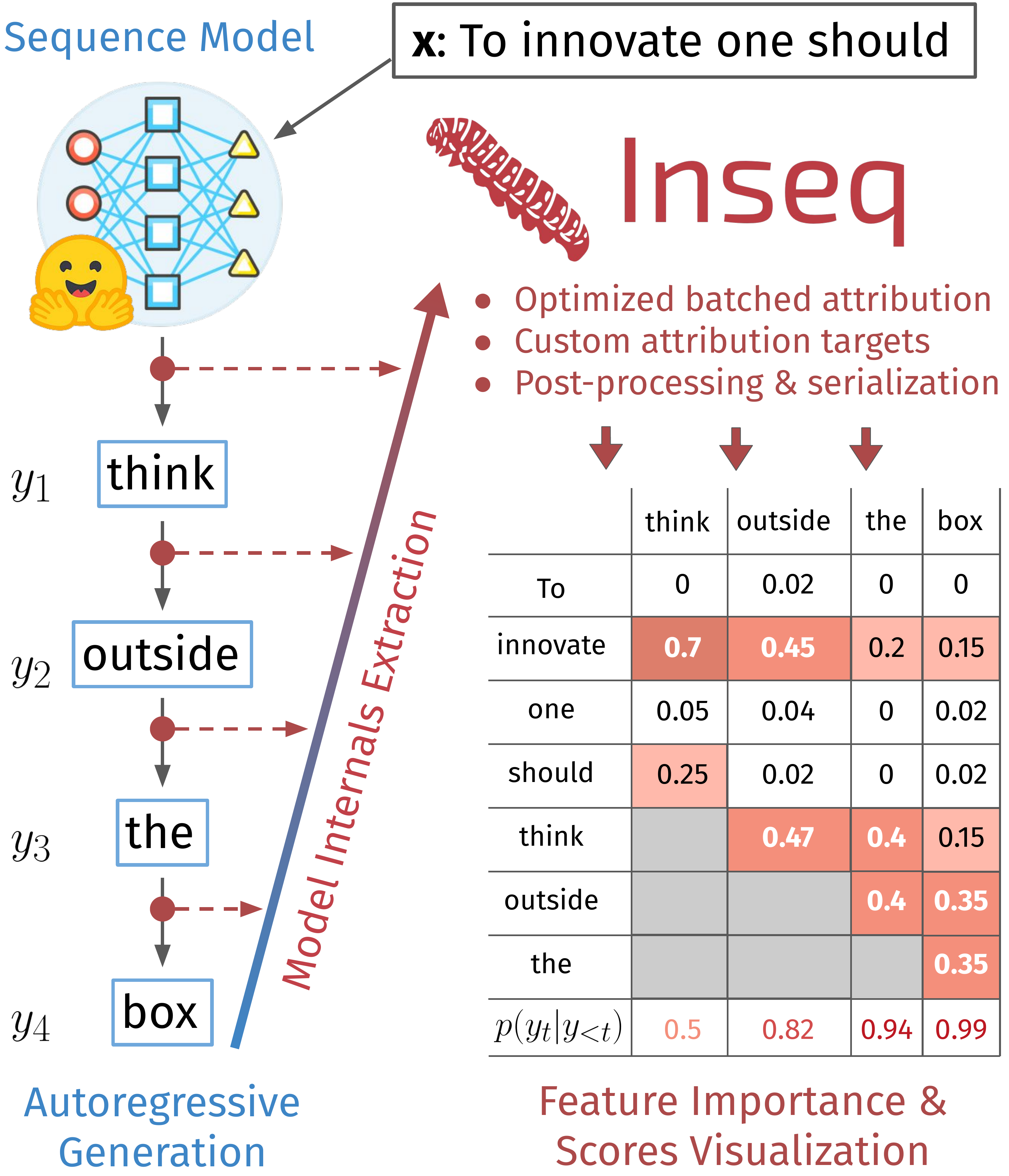}
    \caption{Feature importance and next-step probability extraction and visualization using Inseq with a \hf~Transformers causal language model.} 
    \label{fig:inseq}
\end{figure}
However, feature attribution techniques have mainly been applied to classification settings~\citep{atanasova-etal-2020-diagnostic, wallace-2020-interpreting, madsen-etal-2022-evaluating,chrysostomou-aletras-2022-empirical}, with relatively little interest in the more convoluted mechanisms underlying generation.
Classification attribution is a single-step process resulting in one importance score per input token, often allowing for intuitive interpretations in relation to the predicted class.
Sequential attribution\footnote{We use \textit{sequence generation} to refer to all iterative tasks including (but not limited to) natural language generation.} instead involves a computationally expensive multi-step iteration producing a matrix $A_{ij}$ representing the importance of every input $i$ in the prediction of every generation outcome $j$ (Figure~\ref{fig:inseq}).
Moreover, since previous generation steps causally influence following predictions, they must be dynamically incorporated into the set of attributed inputs throughout the process.
Lastly, while classification usually involves a limited set of classes and simple output selection (e.g. argmax after softmax), generation routinely works with large vocabularies and non-trivial decoding strategies~\citep{eikema-aziz-2020-map}.
These differences limited the use of feature attribution methods for generation settings, with relatively few works improving attribution efficiency~\citep{vafa-etal-2021-rationales,ferrando-etal-2022-towards} and explanations' informativeness~\citep{yin-neubig-2022-interpreting}.

\begin{figure}[!t]
    \centering
    \begin{lstlisting}[language=Python]
import inseq

# Load HF Hub model and attribution method
model = inseq.load_model(
    "google/flan-t5-base",
    "integrated_gradients"
)
# Answer and attribute generation steps
attr_out = model.attribute(
    "Does 3 + 3 equal 6?",
    attribute_target=True
)
# Visualize the generated attribution,
# applying default token-level aggregation
attr_out.show()
\end{lstlisting}
    \includegraphics[width=.75\linewidth]{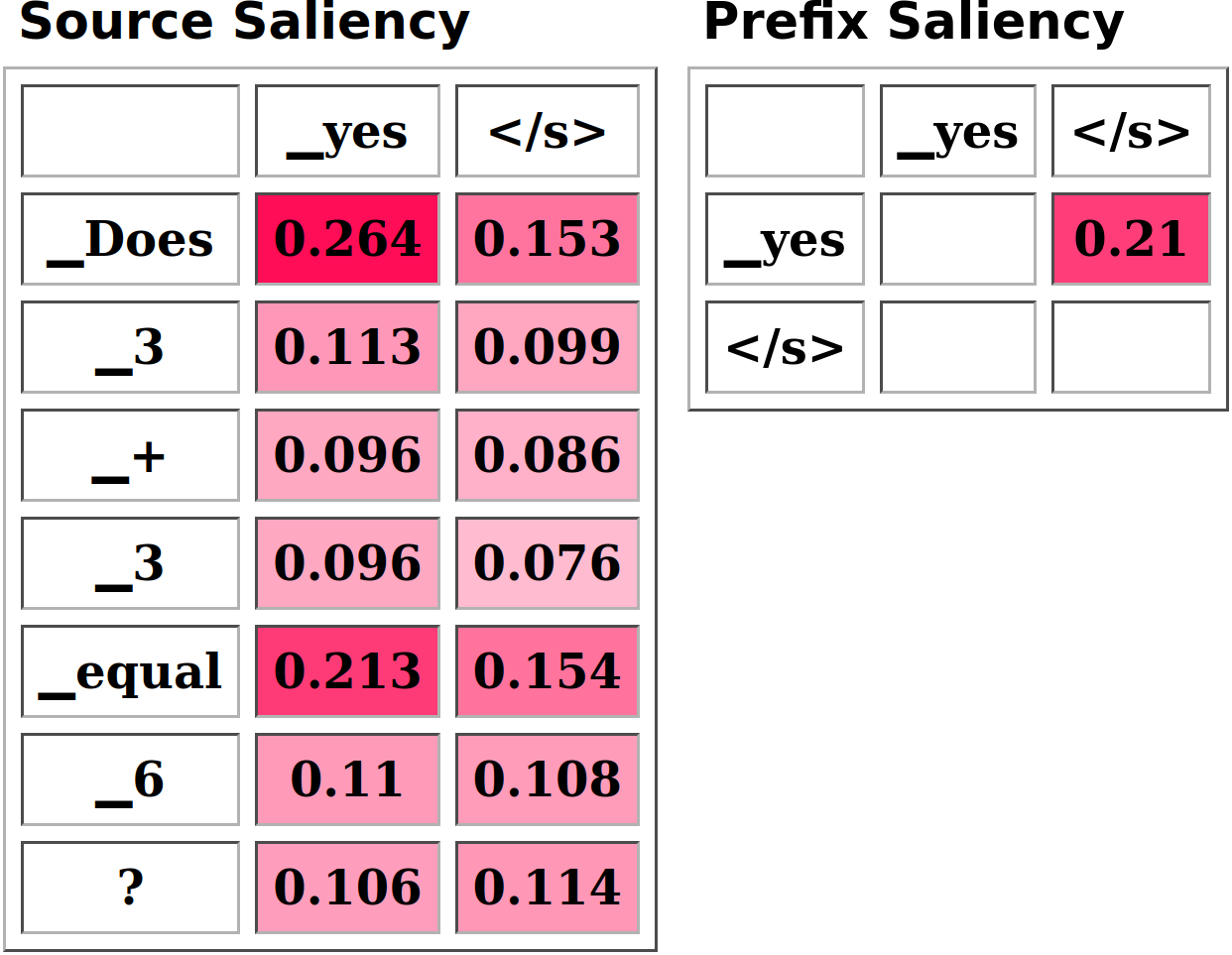}
    \caption{Computing and visualizing source and target-side attributions using Flan-T5~\citep{chung-etal-2022-scaling}.}
    \label{fig:code-short}
\end{figure}

In this work, we introduce \textbf{Inseq}, a Python library to democratize access to interpretability analyses of generative language models.
Inseq centralizes access to a broad set of feature attribution methods, sourced in part from the Captum~\citep{kokhlikyan-etal-2020-captum} framework, enabling a fair comparison of different techniques for all sequence-to-sequence and decoder-only models in the popular \hf~Transformers library~\citep{wolf-etal-2020-transformers}.
Thanks to its intuitive interface, users can easily integrate interpretability analyses into sequence generation experiments with just 3 lines of code (Figure~\ref{fig:code-short}).
Nevertheless, Inseq is also highly flexible, including cutting-edge attribution methods with built-in post-processing features (\cref{sec:feat-attr}), supporting customizable attribution targets and enabling constrained decoding of arbitrary sequences (\cref{sec:customize}).
In terms of usability, Inseq greatly simplifies access to local and global explanations with built-in support for a command line interface (CLI), optimized batching enabling dataset-wide attribution, and various methods to visualize, serialize and reload attribution outcomes and generated sequences (\cref{sec:usability}).
Ultimately, Inseq's aims to make sequence models first-class citizens in interpretability research and drive future advances in interpretability for generative applications.

\section{Related Work}

\paragraph{Feature Attribution for Sequence Generation} Work on feature attribution for sequence generation has mainly focused on machine translation (MT). \citet{bahdanau-etal-2015-neural} showed how attention weights of neural MT models encode interpretable alignment patterns. \citet{alvarez-melis-jaakkola-2017-causal} adopted a perturbation-based framework to highlight biases in MT systems. \citet{ding-etal-2019-saliency,he-etal-2019-towards,voita-etal-2021-analyzing,voita-etal-2021-language} \textit{inter alia} conducted analyses on MT word alignments, coreference resolution and training dynamics with various gradient-based attribution methods. \citet{vafa-etal-2021-rationales,ferrando-etal-2022-towards} developed approaches to efficiently compute sequential feature attributions without sacrificing accuracy. \citet{yin-neubig-2022-interpreting} introduced contrastive feature attribution to disentangle factors influencing generation in language models. Attribution scores obtained from MT models were also used to detect hallucinatory behavior~\citep{dale-etal-2022-detecting,tang-etal-2022-reducing,xu-etal-2023-understanding}, providing a compelling practical use case for such explanations.

\paragraph{Tools for NLP Interpretability} Although many post-hoc interpretability libraries were released recently, only a few support sequential feature attribution. Notably, LIT~\citep{tenney-etal-2020-language}, a structured framework for analyzing models across modalities, and Ecco~\citep{alammar-2021-ecco}, a library specialized in interactive visualizations of model internals. LIT is an all-in-one GUI-based tool to analyze model behaviors on entire datasets. However, the library does not provide out-of-the-box support for \hf~Transformers models, requiring the definition of custom wrappers to ensure compatibility. Moreover, it has a steep learning curve due to its advanced UI, which might be inconvenient when working on a small amount of examples. All these factors limit LIT usability for researchers working with custom models, needing access to extracted scores, or being less familiar with interpretability research. On the other hand, Ecco is closer to our work, being based on \hf~Transformers and having started to support encoder-decoder models concurrently with Inseq development. Despite a marginal overlap in their functionalities, the two libraries provide orthogonal benefits: Inseq's flexible interface makes it especially suitable for methodical quantitative analyses involving repeated evaluations, while Ecco excels in qualitative analyses aimed at visualizing model internals. Other popular tools such as ERASER~\citep{deyoung-etal-2020-eraser}, Thermostat~\citep{feldhus-etal-2021-thermostat}, transformers-interpret~\citep{pierse-2021-transformers} and ferret \citep{attanasio-2022-ferret} do not support sequence models.

\section{Design}

Inseq combines sequence models sourced from \hf~Transformers~\citep{wolf-etal-2020-transformers} and attribution methods mainly sourced from Captum~\citep{kokhlikyan-etal-2020-captum}. While only text-based tasks are currently supported, the library's modular design\footnote{More details are available in Appendix ~\ref{app:design}.} would enable the inclusion of other modeling frameworks (e.g. fairseq~\citep{ott-etal-2019-fairseq}) and modalities (e.g. speech) without requiring substantial redesign. Optional dependencies include \hf~Datasets~\citep{lhoest-etal-2021-datasets} and Rich\footnote{\url{https://github.com/Textualize/rich}}. 

\subsection{Guiding Principles}

\paragraph{Research and Generation-oriented} Inseq should support interpretability analyses of a broad set of sequence generation models without focusing narrowly on specific architectures or tasks. Moreover, the inclusion of new, cutting-edge methods should be prioritized to enable fair comparisons with well-established ones.

\paragraph{Scalable} The library should provide an optimized interface to a wide range of use cases, models and setups, ranging from interactive attributions of individual examples using toy models to compiling statistics of large language models' predictions for entire datasets.

\paragraph{Beginner-friendly} Inseq should provide built-in access to popular frameworks for sequence generation modeling and be fully usable by non-experts at a high level of abstraction, providing sensible defaults for supported attribution methods.

\paragraph{Extensible} Inseq should support a high degree of customization for experienced users, with out-of-the-box support for user-defined solutions to enable future investigations into models' behaviors.

\section{Modules and Functionalities}

\subsection{Feature Attribution and Post-processing}
\label{sec:feat-attr}

\begin{table}
\small
\centering
\begin{tabular}{p{0.2em}llc}
\toprule
& \textbf{Method} & \textbf{Source} & $f(l)$ \\
\midrule
\multirow{5}{*}{\textbf{G}} & (Input $\times$) Gradient & \citeauthor{simonyan-2013-saliency} & \yes \\
& DeepLIFT & \citeauthor{shrikumar-2017-deeplift} & \yes \\
& GradientSHAP & \citeauthor{lundberg-lee-2017-shap} & \no \\
& Integrated Gradients & \citeauthor{sundararajan-etal-2017-ig} & \yes \\
& Discretized IG & \citeauthor{sanyal-ren-2021-discretized} & \no \\
\midrule
\textbf{I} & Attention Weights & \citeauthor{bahdanau-etal-2015-neural} & \yes \\
\midrule
\multirow{3}{*}{\textbf{P}} & Occlusion (Blank-out) &  \citeauthor{zeiler-fergus-2014-visualizing} & \no \\
& LIME & \citeauthor{ribeiro-2016-lime} & \no \\
\midrule
\midrule
\multirow{4}{*}{\textbf{S}} & (Log) Probability & - \\
& Softmax Entropy & - \\
& Target Cross-entropy & - \\
& Perplexity & - \\
& Contrastive Prob. $\Delta$ & \citeauthor{yin-neubig-2022-interpreting} \\
& $\mu$ MC Dropout Prob. & \citeauthor{gal-ghahramani-2016-dropout} \\
\bottomrule
\end{tabular}
\caption{Overview of gradient-based (\textbf{G}), internals-based (\textbf{I}) and perturbation-based (\textbf{P}) attribution methods and built-in step functions (\textbf{S}) available in Inseq. $f(l)$ marks methods allowing for attribution of arbitrary intermediate layers.}
\label{tab:methods}
\end{table}

At its core, Inseq provides a simple interface to apply feature attribution techniques for sequence generation tasks. We categorize methods in three groups, \textit{gradient-based}, \textit{internals-based} and \textit{perturbation-based}, depending on their underlying approach to importance quantification.\footnote{We distinguish between gradient- and internals-based methods to account for their difference in scores' granularity.} Table~\ref{tab:methods} presents the full list of supported methods. Aside from popular model-agnostic methods, Inseq notably provides built-in support for attention weight attribution and the cutting-edge Discretized Integrated Gradients method~\citep{sanyal-ren-2021-discretized}. Moreover, multiple methods allow for the importance attribution of custom intermediate model layers, simplifying studies on representational structures and information mixing in sequential models, such as our case study of Section~\ref{sec:rome-repro}.

\paragraph{Source and target-side attribution} When using encoder-decoder architectures, users can set the {\small\texttt{attribute\_target}} parameter to include or exclude the generated prefix in the attributed inputs. In most cases, this should be desirable to account for recently generated tokens when explaining model behaviors, such as when to terminate the generation (e.g. relying on the presence {\small\texttt{\_yes}} in the target prefix to predict {\small\texttt{</s>}} in Figure~\ref{fig:code-short}, bottom-right matrix). However, attributing the source side separately could prove useful, for example, to derive word alignments from importance scores.

\paragraph{Post-processing of attribution outputs} Aggregation is a fundamental but often overlooked step in attribution-based analyses since most methods produce neuron-level or subword-level importance scores that would otherwise be difficult to interpret. Inseq includes several {\small\texttt{Aggregator}} classes to perform attribution aggregation across various dimensions. For example, the input word ``Explanation'' could be tokenized in two subword tokens ``Expl'' and ``anation'', and each token would receive $N$ importance scores, with $N$ being the model embedding dimension. In this case, aggregators could first merge subword-level scores into word-level scores, and then merge granular embedding-level scores to obtain a single token-level score that is easier to interpret. Moreover, aggregation could prove especially helpful for long-form generation tasks such as summarization, where word-level importance scores could be aggregated to obtain a measure of sentence-level relevance. Notably, Inseq allows chaining multiple aggregators like in the example above using the {\small\texttt{AggregatorPipeline}} class, and provides a {\small \texttt{PairAggregator}} to aggregate different attribution maps, simplifying the conduction of contrastive analyses as in Section~\ref{sec:gender-bias}.\footnote{See Appendix~\ref{app:pair-agg-gender-swap} for an example.}

\subsection{Customizing generation and attribution}
\label{sec:customize}

During attribution, Inseq first generates target tokens using \hf~Transformers and then attributes them step by step. If a custom target string is specified alongside model inputs, the generation step is instead skipped, and the provided text is attributed by constraining the decoding of its tokens\footnote{Constrained decoding users should be aware of its limitations in the presence of a high distributional discrepancy with natural model outputs~\citep{vamvas-sennrich-2021-limits}.}. Constrained attribution can be used, among other things, for contrastive comparisons of minimal pairs and to obtain model justifications for desired outputs.
\paragraph{Custom step functions} At every attribution step, Inseq can use models' internal information to extract scores of interest (e.g. probabilities, entropy) that can be useful, among other things, to quantify model uncertainty (e.g. how likely the generated {\small\texttt{\_yes}} token was given the context in Figure~\ref{fig:code-short}). Inseq provides access to multiple built-in step functions (Table~\ref{tab:methods}, \textbf{S}) enabling the computation of these scores, and allows users to create and register new custom ones. Step scores are computed together with the attribution, returned as separate sequences in the output, and visualized alongside importance scores (e.g. the $p(y_t|y_{<t})$ row in Figure~\ref{fig:inseq}).
\paragraph{Step functions as attribution targets} For methods relying on model outputs to predict input importance (gradient and perturbation-based), feature attributions are commonly obtained from the model's output logits or class probabilities \cite{bastings-etal-2022-will}. However, recent work showed the effectiveness of using targets such as the probability difference of a contrastive output pair to answer interesting questions like ``What inputs drive the prediction of $y$ rather than $\hat{y}$?'' \cite{yin-neubig-2022-interpreting}. In light of these advances, Inseq users can leverage any built-in or custom-defined step function as an attribution target, enabling advanced use cases like contrastive comparisons and uncertainty-weighted attribution using MC Dropout~\citep{gal-ghahramani-2016-dropout}.

\subsection{Usability Features}
\label{sec:usability}

\paragraph{Batched and span-focused attributions} The library provides built-in batching capabilities, enabling users to go beyond single sentences and attribute even entire datasets in a single function call. When the attribution of a specific span of interest is needed, Inseq also allows specifying a start and end position for the attribution process. This functionality greatly accelerates the attribution process for studies on localized phenomena (e.g. pronoun coreference in MT models).

\paragraph{CLI, Serialization and Visualization} The Inseq library offers an API to attribute single examples or entire \hf~Datasets from the command line and save resulting outputs and visualizations to a file. Attribution outputs can be saved and loaded in JSON format with their respective metadata to easily identify the provenance of contents. Attributions can be visualized in the command line or IPython notebooks and exported as HTML files.

\paragraph{Quantized Model Attribution} Supporting the attribution of large models is critical given recent scaling tendencies~\citep{kaplan-etal-2020-scaling}. All models allowing for quantization using {\small \texttt{bitsandbytes}}~\citep{dettmers-etal-2022-gpt3} can be loaded in 8-bit directly from \hf~Transformers, and their attributions can be computed normally using Inseq.\footnote{{\scriptsize \texttt{bitsandbytes 0.37.0}} required for backward method, see Appendix~\ref{app:ex-factual-int8} for an example.} A minimal manual evaluation of 8-bit attribution outputs for Section~\ref{sec:rome-repro} study shows minimal discrepancies compared to full-precision results.

\section{Case Studies}

\subsection{Gender Bias in Machine Translation}
\label{sec:gender-bias}
In the first case study, we use Inseq to investigate gender bias in MT models.
Studying social biases embedded in these models is crucial to understand and mitigate the representational and allocative harms they might engender~\cite{blodgett-etal-2020-language}.
\citet{savoldi-etal-2021-gender} note that the study of bias in MT could benefit from explainability techniques to identify spurious cues exploited by the model and the interaction of different features that can lead to intersectional bias.

\paragraph{Synthetic Setup: Turkish to English}
\label{sec:gender-bias-turkish}

The Turkish language uses the gender-neutral pronoun \textit{o}, which can be translated into English as either ``he'', ``she'', or ``it'', making it interesting to study gender bias in MT when associated with a language such as English for which models will tend to choose a gendered pronoun form.
Previous works leveraged translations from gender-neutral languages to show gender bias present in translation systems~\cite{cho-etal-2019-measuring,prates2020assessing,farkas2022measure}.
We repeat this simple setup using a Turkish-to-English MarianMT model~\citep{tiedemann-2020-tatoeba} and compute different metrics to quantify gender bias using Inseq.

We select 49 Turkish occupation terms verified by a native speaker (see Appendix \ref{app:gender-bias-turkish}) and use them to infill the template sentence ``\textit{O bir} \underline{\hspace{0.4cm}}'' (He/She is a(n) \underline{\hspace{0.4cm}}).
For each translation, we compute attribution scores for source Turkish pronoun ($x_\text{pron}$) and occupation ($x_\text{occ}$) tokens\footnote{For multi-token occupation terms, e.g., \textit{bilim insanı} (scientist), the attribution score of the first token was used.} when generating the target English pronoun ($y_\text{pron}$) using Integrated Gradients (IG), Gradients ($\nabla$), and Input $\times$ Gradient (I$\times$G),\footnote{We set approx. steps to ensure convergence $\Delta < 0.05$ for IG. All methods use the L2 norm to obtain token-level attributions.}. We also collect target pronoun probabilities ($p(y_\text{pron})$), rank the 49 occupation terms using these metrics, and finally compute Kendall's $\tau$ correlation with the percentage of women working in the respective fields, using U.S. labor statistics as in previous works~\cite[e.g.,][]{caliskan2017semantics,rudinger-etal-2018-gender}.
Table~\ref{tab:turkish_gender_bias} presents our results.

In the \textbf{base case}, we correlate the different metrics with how much the gender distribution deviates from an equal distribution ($50-50\%$) for each occupation (i.e., the gender bias irrespective of the direction).
We observe a strong gender bias, with ``she'' being chosen only for 5 out of 49 translations and gender-neutral variants never being produced by the MT model.
We find a low correlation between pronoun probability and the degree of gender stereotype associated with the occupation. Moreover, we note a weaker correlation for IG compared to the other two methods. For those, attribution scores for $x_\text{occ}$ show significant correlations with labor statistics, supporting the intuition that the MT model will accord higher importance to source occupation terms associated to gender-stereotypical occupations when predicting the gendered target pronoun.

In the \textbf{gender-swap case} (\female~$\rightarrow$~\male), we use the {\small \texttt{PairAggregator}} class to contrastively compare attribution scores and probabilities when translating the pronoun as ``She'' or ``He''.\footnote{An example is provided in Appendix~\ref{app:pair-agg-gender-swap}.}
We correlate resulting scores with the \% of women working in the respective occupation and find strong correlations for $p(y_\text{pron})$, supporting the validity of contrastive approaches in uncovering gender bias.

\begin{table}
\centering
\small
\begin{tabular}{crr|rr}
\toprule
 & \multicolumn{2}{c}{Base} & \multicolumn{2}{c}{\female~$\rightarrow$~\male}\\
 \cmidrule{2-5}
  & $x_\text{pron}$ & $x_\text{occ}$ & $x_\text{pron}$ & $x_\text{occ}$ \\
\midrule
$p(y_\text{pron})$ & \multicolumn{2}{c}{{\cellcolor[HTML]{FDFEBC}} 0.01\notsig} & \multicolumn{2}{c}{{\cellcolor[HTML]{FBA05B}} -0.44\sig} \\
\midrule
$\nabla$ & {\cellcolor[HTML]{FEE695}}  -0.16\notsig & {\cellcolor[HTML]{CBE982}}  0.25\sig  &{\cellcolor[HTML]{D1EC86}}  0.23\sig & {\cellcolor[HTML]{FFFEBE}}  -0.00\notsig\\
IG & {\cellcolor[HTML]{FFF2AA}}  -0.08\notsig & {\cellcolor[HTML]{ECF7A6}}  0.09\notsig & {\cellcolor[HTML]{EBF7A3}}  0.11\notsig & {\cellcolor[HTML]{DFF293}}  0.17\notsig\\
I$\times$G & {\cellcolor[HTML]{FEEDA1}}  -0.11\notsig & {\cellcolor[HTML]{D3EC87}}  0.22\sig &{\cellcolor[HTML]{D5ED88}}  0.22\sig & {\cellcolor[HTML]{FFFDBC}}  -0.01\notsig\\
\bottomrule
\end{tabular}
\caption{\textbf{Gender Bias in Turkish-to-English MT:} Kendall's $\tau$ correlation of MT model metrics with U.S. labor statistics. $*$ = Significant correlation ($p<.05$).}
\label{tab:turkish_gender_bias}
\end{table}

\paragraph{Qualitative Example: English to Dutch}
\label{sec:gender-bias-bug}

We qualitatively analyze biased MT outputs, showing how attributions can help develop hypotheses about models' behavior.
Table \ref{tab:m2m-gender-example} (top) shows the I~$\times$~G attributions for English-to-Dutch translation using M2M-100~\citep[418M,][]{fan-etal-2021-m2m100}. The model mistranslates the pronoun ``her'' into the masculine form \textit{zijn} (his). 
We find that the wrongly translated pronoun exhibits high probability but does not associate substantial importance to the source occupation term ``teacher''.
Instead, we find good relative importance for the preceding word and \textit{leraar} (male teacher). This suggests a strong prior bias for masculine variants, shown by the pronoun \textit{zijn} and the noun \textit{leraar}, as a possible cause for this mistranslation.
When considering the contrastive example obtained by swapping \textit{leraar} with its gender-neutral variant \textit{leerkracht} (Table~\ref{tab:m2m-gender-example}, bottom), we find increased importance of the target occupation in determining the correctly-gendered target pronoun \textit{haar} (her). 
Our results highlight the tendency of MT models to attend inputs sequentially rather than relying on context, hinting at the known benefits of context-aware models for pronoun translation~\citep{voita-etal-2018-context}.

\begin{table}
\scriptsize
    \begin{tabular}{L{1cm}C{0.7cm}C{0.8cm}C{0.9cm}C{0.7cm}C{0.7cm}} %{lrrrrr}%{
\toprule
\textbf{Source} & De & leraar & verliest & zijn & baan \\
\midrule
The & {\cellcolor[HTML]{FFE4E4}} \color[HTML]{000000} 0.10 & {\cellcolor[HTML]{FFECEC}} \color[HTML]{000000} 0.08 & {\cellcolor[HTML]{FFF4F4}} \color[HTML]{000000} 0.04 & {\cellcolor[HTML]{FFF8F8}} \color[HTML]{000000} 0.03 & {\cellcolor[HTML]{FFFAFA}} \color[HTML]{000000} 0.02 \\
teacher & {\cellcolor[HTML]{FFE2E2}} \color[HTML]{000000} 0.11 & {\cellcolor[HTML]{FFCACA}} \color[HTML]{000000} 0.20 & {\cellcolor[HTML]{FFF0F0}} \color[HTML]{000000} 0.06 & {\cellcolor[HTML]{FFF6F6}} \color[HTML]{000000} 0.03 & {\cellcolor[HTML]{FFF2F2}} \color[HTML]{000000} 0.05 \\
loses & {\cellcolor[HTML]{FFE2E2}} \color[HTML]{000000} 0.11 & {\cellcolor[HTML]{FFE8E8}} \color[HTML]{000000} 0.09 & {\cellcolor[HTML]{FFC0C0}} \color[HTML]{000000} 0.25 & {\cellcolor[HTML]{FFEEEE}} \color[HTML]{000000} 0.07 & {\cellcolor[HTML]{FFECEC}} \color[HTML]{000000} 0.07 \\
her & {\cellcolor[HTML]{FFD8D8}} \color[HTML]{000000} 0.15 & {\cellcolor[HTML]{FFE8E8}} \color[HTML]{000000} 0.09 & {\cellcolor[HTML]{FFE6E6}} \color[HTML]{000000} 0.10 & {\cellcolor[HTML]{FFC8C8}} \color[HTML]{000000} 0.21 & {\cellcolor[HTML]{FFEEEE}} \color[HTML]{000000} 0.07 \\
job & {\cellcolor[HTML]{FFE6E6}} \color[HTML]{000000} 0.10 & {\cellcolor[HTML]{FFECEC}} \color[HTML]{000000} 0.08 & {\cellcolor[HTML]{FFEAEA}} \color[HTML]{000000} 0.08 & {\cellcolor[HTML]{FFE4E4}} \color[HTML]{000000} 0.10 & {\cellcolor[HTML]{FFC2C2}} \color[HTML]{000000} 0.24 \\
\midrule
\textbf{Target} & De & leraar & verliest & zijn & baan \\
\midrule
De & { \color[HTML]{F1F1F1}}  & {\cellcolor[HTML]{FFC6C6}} \color[HTML]{000000} 0.23 & {\cellcolor[HTML]{FFF2F2}} \color[HTML]{000000} 0.05 & {\cellcolor[HTML]{FFF0F0}} \color[HTML]{000000} 0.06 & {\cellcolor[HTML]{FFF6F6}} \color[HTML]{000000} 0.04 \\
leraar & { \color[HTML]{F1F1F1}}  & { \color[HTML]{F1F1F1}}  & {\cellcolor[HTML]{FFD4D4}} \color[HTML]{000000} 0.17 & {\cellcolor[HTML]{FFDEDE}} \color[HTML]{000000} 0.13 & {\cellcolor[HTML]{FFF6F6}} \color[HTML]{000000} 0.03 \\
verliest & { \color[HTML]{F1F1F1}}  & { \color[HTML]{F1F1F1}}  & { \color[HTML]{F1F1F1}}  & {\cellcolor[HTML]{FFD0D0}} \color[HTML]{000000} 0.18 & {\cellcolor[HTML]{FFECEC}} \color[HTML]{000000} 0.08 \\
zijn & { \color[HTML]{F1F1F1}}  & { \color[HTML]{F1F1F1}}  & { \color[HTML]{F1F1F1}}  & { \color[HTML]{F1F1F1}}  & {\cellcolor[HTML]{FFBEBE}} \color[HTML]{000000} 0.26 \\
\midrule
$p(y_t)$ &  0.69 &  0.28 &  0.35 &  0.65 &  0.29 \\
\bottomrule
\end{tabular}
% \hfill
\vspace{0.1cm}\\
\begin{tabular}{L{1cm}C{0.7cm}C{0.8cm}C{0.9cm}C{0.7cm}C{0.7cm}}
\toprule
\textbf{Source} & De & \male → $\circ$ & verliest & haar & baan \\
\midrule
The & {\cellcolor[HTML]{FFFEFE}} \color[HTML]{000000} 0.00 & {\cellcolor[HTML]{F6F6FF}} \color[HTML]{000000} -0.02 & {\cellcolor[HTML]{FFFEFE}} \color[HTML]{000000} 0.00 & {\cellcolor[HTML]{FFFEFE}} \color[HTML]{000000} 0.00 & {\cellcolor[HTML]{FFFEFE}} \color[HTML]{000000} 0.00 \\
teacher & {\cellcolor[HTML]{FFFEFE}} \color[HTML]{000000} 0.00 & {\cellcolor[HTML]{EEEEFF}} \color[HTML]{000000} -0.05 & {\cellcolor[HTML]{FFFCFC}} \color[HTML]{000000} -0.01 & {\cellcolor[HTML]{FFFCFC}} \color[HTML]{000000} -0.01 & {\cellcolor[HTML]{FFFCFC}} \color[HTML]{000000} -0.01 \\
loses & {\cellcolor[HTML]{FFFEFE}} \color[HTML]{000000} 0.00 & {\cellcolor[HTML]{F6F6FF}} \color[HTML]{000000} -0.02 & {\cellcolor[HTML]{FFFCFC}} \color[HTML]{000000} -0.01 & {\cellcolor[HTML]{F6F6FF}} \color[HTML]{000000} -0.02 & {\cellcolor[HTML]{FFFCFC}} \color[HTML]{000000} -0.01 \\
her & {\cellcolor[HTML]{FFFEFE}} \color[HTML]{000000} 0.00 & {\cellcolor[HTML]{FFFCFC}} \color[HTML]{000000} -0.01 & {\cellcolor[HTML]{FFFCFC}} \color[HTML]{000000} -0.01 & {\cellcolor[HTML]{D8D8FF}} \color[HTML]{000000} -0.10 & {\cellcolor[HTML]{FFFAFA}} \color[HTML]{000000} 0.01 \\
job & {\cellcolor[HTML]{FFFEFE}} \color[HTML]{000000} 0.00 & {\cellcolor[HTML]{F6F6FF}} \color[HTML]{000000} -0.02 & {\cellcolor[HTML]{FFFCFC}} \color[HTML]{000000} -0.01 & {\cellcolor[HTML]{F6F6FF}} \color[HTML]{000000} -0.02 & {\cellcolor[HTML]{F6F6FF}} \color[HTML]{000000} -0.02 \\
\midrule
\textbf{Target} & De & \male → $\circ$ & verliest & haar & baan \\
\midrule
De & \color[HTML]{F1F1F1}  & {\cellcolor[HTML]{EEEEFF}} \color[HTML]{000000} -0.07 & {\cellcolor[HTML]{FCFCFF}} \color[HTML]{000000} -0.01 & {\cellcolor[HTML]{FFFCFC}} \color[HTML]{000000} 0.01 & {\cellcolor[HTML]{FCFCFF}} \color[HTML]{000000} -0.01 \\
\male → $\circ$ &  \color[HTML]{F1F1F1}  &  \color[HTML]{F1F1F1}  & {\cellcolor[HTML]{FFE8E8}} \color[HTML]{000000} 0.09 & {\cellcolor[HTML]{FFD0D0}} \color[HTML]{000000} 0.18 & {\cellcolor[HTML]{FFFAFA}} \color[HTML]{000000} 0.02 \\
verliest &  \color[HTML]{F1F1F1}  &  \color[HTML]{F1F1F1}  &  \color[HTML]{F1F1F1}  & {\cellcolor[HTML]{F6F6FF}} \color[HTML]{000000} -0.03 & {\cellcolor[HTML]{FEFEFF}} \color[HTML]{000000} 0.00 \\
haar &  \color[HTML]{F1F1F1}  &  \color[HTML]{F1F1F1}  &  \color[HTML]{F1F1F1}  &  \color[HTML]{F1F1F1}  & {\cellcolor[HTML]{FFFEFE}} \color[HTML]{000000} 0.00 \\
\midrule
$\Delta p(y_t)$ & 0.00 &  -0.23 &  0.13 &  0.20 &  0.00 \\
\bottomrule
\end{tabular}
    \caption{\textbf{Top:} Attribution of pronoun gender mistranslation using M2M-100. \textbf{Bottom:} Target attribution difference when swapping the target noun gender (\male → $\circ$) from \textit{leraar} (male) to \textit{leerkracht} (gender-neutral).}
    \label{tab:m2m-gender-example}
\end{table}

\subsection{Locating Factual Knowledge inside GPT-2 with Contrastive Attribution Tracing}
\label{sec:rome-repro}

\begin{figure}
    \centering
    \includegraphics[width=\linewidth, angle=0]{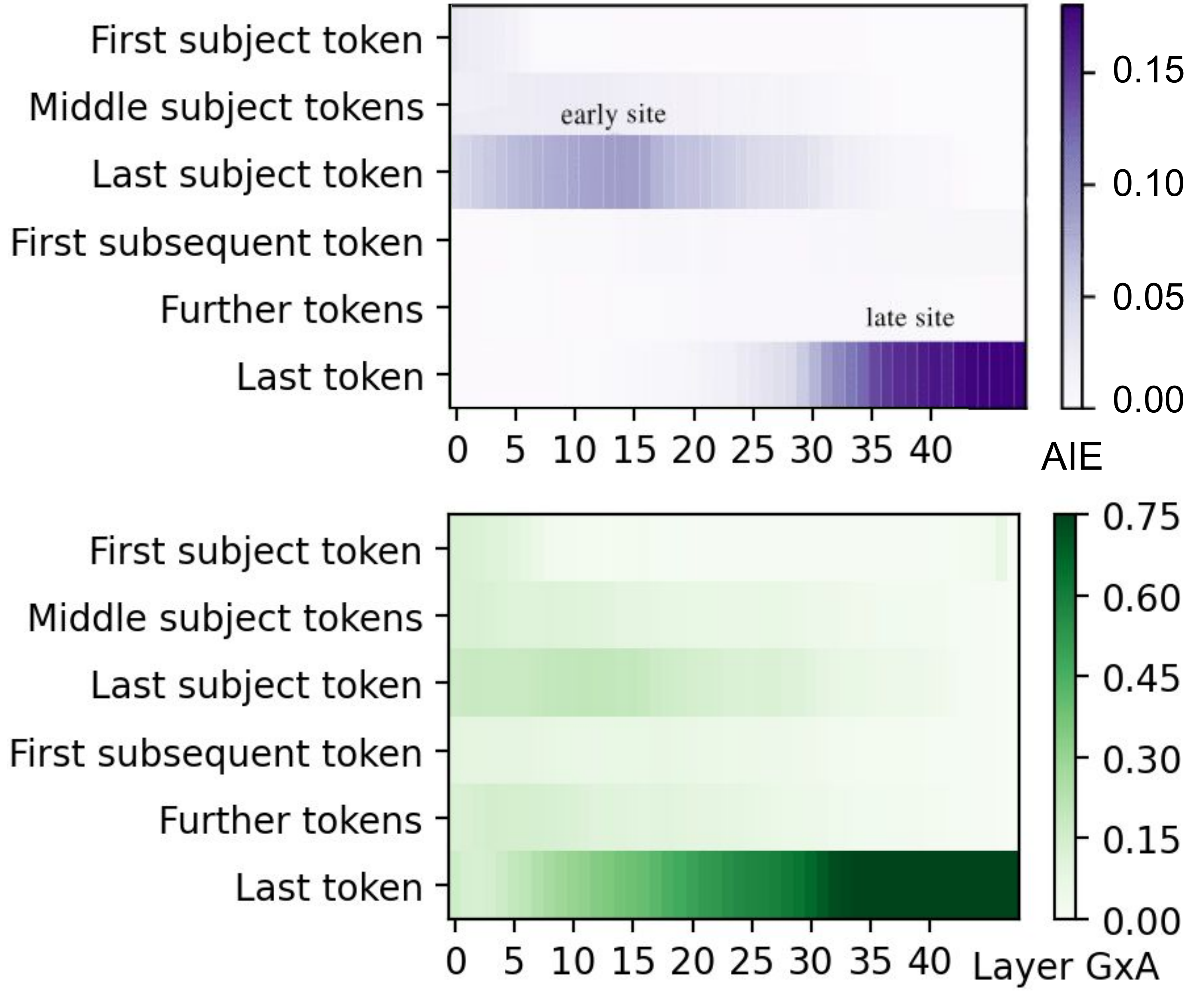}
    \caption{\textbf{Top:} Estimated causal importance of GPT-2 XL layers for predicting factual associations, as reported by \citet{meng-2022-rome}. \textbf{Bottom:} Average GPT-2 XL Gradient $\times$ Layer Activation scores obtained with Inseq using contrastive factual pairs as attribution targets.} 
    \label{fig:rome-repro}
\end{figure}

For our second case study, we experiment with a novel attribution-based technique to locate factual knowledge encoded in the layers of GPT-2 1.5B~\citep{radford-etal-2019-language}. Specifically, we aim to reproduce the results of~\citet{meng-2022-rome}, showing the influence of intermediate layers in mediating the recall of factual statements such as \textit{`The Eiffel Tower is located in the city of} $\rightarrow$ \textit{Paris'}. \citet{meng-2022-rome} estimate the effect of network components in the prediction of factual statements as the difference in probability of a correct target (e.g. \textit{Paris}), given a corrupted subject embedding (e.g. for \textit{Eiffel Tower}), before and after restoring clean activations for some input tokens at different layers of the network. Apart from the obvious importance of final token states in terminal layers, their results highlight the presence of an early site associated with the last subject token playing an important role in recalling the network's factual knowledge (Figure~\ref{fig:rome-repro}, top).

To verify such results, we propose a novel knowledge location method, which we name Contrastive Attribution Tracing (CAT), adopting the contrastive attribution paradigm of~\citet{yin-neubig-2022-interpreting} to locate relevant network components by attributing minimal pairs of correct and wrong factual targets (e.g. \textit{Paris} vs. \textit{Rome} for the example above). To perform the contrastive attribution, we use the Layer Gradient $\times$ Activation method, a layer-specific variant of Input $\times$ Gradient, to propagate gradients up to intermediate network activations instead of reaching input tokens. The resulting attribution scores hence answer the question ``How important are layer $L$ activations for prefix token $t$ in predicting the correct factual target over a wrong one?''. We compute attribution scores for 1000 statements taken from the Counterfact Statement dataset~\citep{meng-2022-rome} and present averaged results in Figure~\ref{fig:rome-repro} (bottom).\footnote{Figure~\ref{fig:ex-factual-knowledge} of Appendix~\ref{app:ex-factual-int8} presents some examples.} Our results closely match those of the original authors, providing further evidence of how attribution methods can be used to identify salient network components and guide model editing, as shown by~\citet{dai-etal-2022-knowledge} and~\citet{nanda-2023-attribution}.

To our best knowledge, the proposed CAT method is the most efficient knowledge location technique to date, requiring only a single forward and backward pass of the attributed model. Patching-based approaches such as causal mediation~\citep{meng-2022-rome}, on the other hand, provide causal guarantees of feature importance at the price of being more computationally intensive. Despite lacking the causal guarantees of such methods, CAT can provide an approximation of feature importance and greatly simplify the study of knowledge encoded in large language model representations thanks to its efficiency.

\section{Conclusion}
\label{sec:conclusion}

We introduced Inseq, an easy-to-use but versatile toolkit for interpreting sequence generation models. With many libraries focused on the study of classification models, Inseq is the first tool explicitly aimed at analyzing systems for tasks such as machine translation, code synthesis, and dialogue generation. Researchers can easily add interpretability evaluations to their studies using our library to identify unwanted biases and interesting phenomena in their models' predictions. We plan to provide continued support and explore developments for Inseq,\footnote{Planned developments available in Appendix \ref{app:next-steps}.} to provide simple and centralized access to a comprehensive set of thoroughly-tested implementations for the interpretability community. In conclusion, we believe that Inseq has the potential to drive real progress in explainable language generation by accelerating the development of new analysis techniques, and we encourage members of this research field to join our development efforts.

%Gabriele: Removed for submission, will be in camera ready
% 
\section*{Acknowledgments}
We thank Ece Takmaz for verifying the Turkish word list used in Section \ref{sec:gender-bias-turkish}.
GS and AB acknowledge the support of the Dutch Research Council (NWO) as part of the project InDeep (NWA.1292.19.399). NF is supported by the German Federal Ministry of Education and Research as part of the project XAINES (01IW20005). OvdW's contributions are financed by the NWO as part of the project ``The biased reality of online media -- Using stereotypes to make media manipulation visible'' (406.DI.19.059).

% Note: The following does NOT count towards the 6-page limit.
\section*{Broader Impact and Ethics Statement}

\paragraph{Reliability of Attribution Methods} The plausibility and faithfulness of attribution methods supported by Inseq is an active matter of debate in the research community, without clear-cut guarantees in identifying specific model behaviors, and prone to users' own biases~\citep{jacovi-goldberg-2020-towards}. We emphasize that explanations produced with Inseq should \underline{not} be adopted in high-risk and user-facing contexts. We encourage Inseq users to critically approach results obtained from our toolkit and validate them on a case-by-case basis.

\paragraph{Technical Limitations and Contributions} While Inseq greatly simplifies comparisons across different attribution methods to ensure their mutual consistency, it does not provide explicit ways of evaluating the quality of produced attributions in terms of faithfulness or plausibility. Moreover, many recent methods still need to be included due to the rapid pace of interpretability research in natural language processing and the small size of our development team. To foster an open and inclusive development environment, we encourage all interested users and new methods' authors to contribute to the development of Inseq by adding their interpretability methods of interest.  

\paragraph{Gender Bias Case Study} The case study of Section~\ref{sec:gender-bias} assumes a simplified concept of binary gender to allow for a more straightforward evaluation of the results. However, we encourage other researchers to consider non-binary gender and different marginalized groups in future bias studies. We acknowledge that measuring bias in language models is complex and that care must be taken in its conceptualization and validation~\cite{blodgett-etal-2020-language,van2022undesirable,bommasani2022trustworthy}, even more so in multilingual settings~\cite{talat-etal-2022-reap}. For this reason, we do not claim to provide a definite bias analysis of these MT models -- especially in light of the aforementioned attributions' faithfulness issues. The study's primary purpose is to demonstrate how attribution methods could be used for exploring social biases in sequence-to-sequence models and showcase the related Inseq functionalities.

\bibliography{anthology,custom}
\bibliographystyle{acl_natbib}

\clearpage
\pagebreak

\appendix

\section{Authors' Contributions}

Authors jointly contributed to the writing and revision of the paper.

\paragraph{Gabriele Sarti} Organized and led the project, developed the first public release of the Inseq library, conducted the case study of Section~\ref{sec:rome-repro}.

\paragraph{Nils Feldhus} Implemented the perturbation-based methods in Inseq and contributed to the  validation of the case study of Section~\ref{sec:rome-repro}.

\paragraph{Ludwig Sickert} Implemented the attention-based attribution method in Inseq.

\paragraph{Oskar van der Wal} Conducted the experiments in the gender bias case study of Section~\ref{sec:gender-bias}.

\paragraph{Malvina Nissim} and \textbf{Arianna Bisazza} ensured the soundness of the overall process and provided valuable inputs for the initial design of the toolkit.

\section{Additional Design Details}
\label{app:design}

\begin{figure*}
    \includegraphics[width=\textwidth, angle=0]{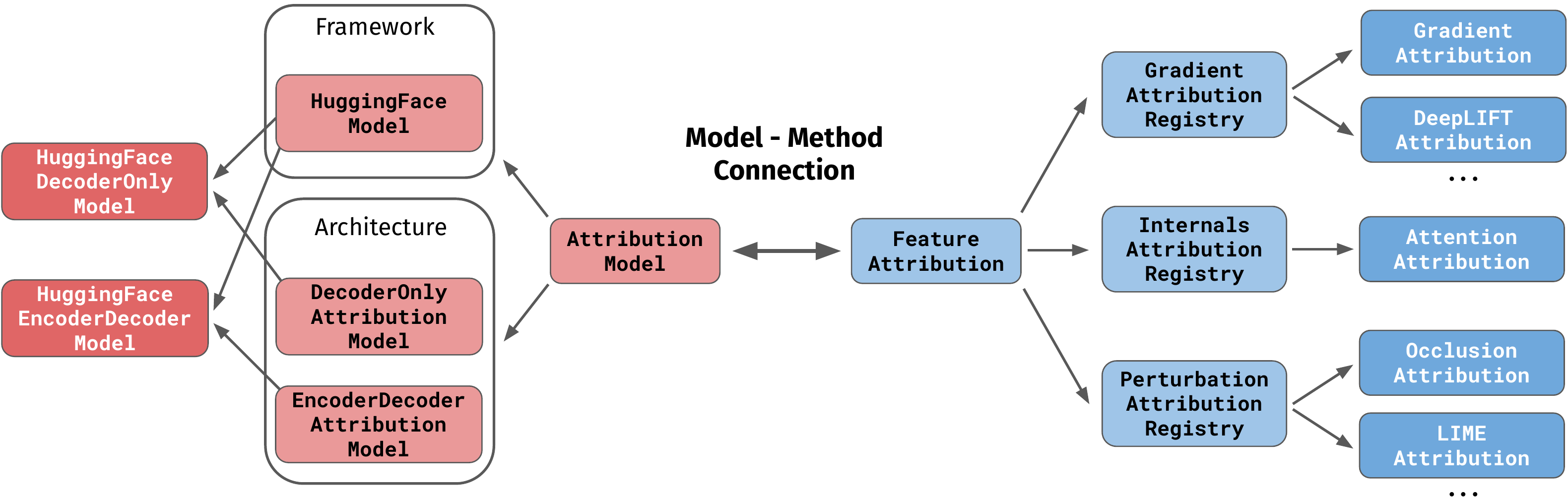}
    \caption{Inseq models and attribution methods. {\small\colorbox{lightreddim}{\textcolor{white}{\textbf{\texttt{Concrete}}}}\colorbox{lightbluedim}{\textcolor{white}{\textbf{\texttt{classes}}}}} combine abstract {\small\colorbox{lightredclear}{\texttt{framework}}} and {\small\colorbox{lightredclear}{\texttt{architecture}}} attribution models classes, and are derived from abstract attribution methods' {\small\colorbox{lightblueclear}{\texttt{categories}}}.} 
    \label{fig:classes}
\end{figure*}

Figure~\ref{fig:classes} presents the Inseq hierarchy of models and attribution methods. The model-method connection enables out-of-the-box attribution using the selected method. Framework-specific and architecture-specific classes enable extending Inseq to new modeling architectures and frameworks. 

\section{Example of Pair Aggregation for Contrastive MT Comparison}
\label{app:pair-agg-gender-swap}

\begin{figure*}[!t]
    \centering
    \begin{lstlisting}[language=Python]
import inseq
from inseq.data.aggregator import AggregatorPipeline, SubwordAggregator, SequenceAttributionAggregator, PairAggregator

# Load the TR-EN translation model and attach the IG method
model = inseq.load_model("Helsinki-NLP/opus-mt-tr-en", "integrated_gradients")

# Batch attribute with forced decoding. Return probabilities, no target attr.
out = model.attribute(
    ["O bir teknisyen", "O bir teknisyen"],
    ["She is a technician.","He is a technician."],
    step_scores=["probability"],
    # The following attributes are specific to the IG method
    internal_batch_size=100,
    n_steps=300
)

# Aggregation pipeline composed by two steps:
# 1. Aggregate subword tokens across all dimensions: [l1, l2, dim] -> [l3, l4, dim]
# 2. Aggregate hidden size to produce token-level attributions: [l1, l2, dim] -> [l1, l2]
subw_aggregator = AggregatorPipeline([SubwordAggregator, SequenceAttributionAggregator])

# Aggregate attributions using the pipeline
masculine = out.sequence_attributions[0].aggregate(aggregator=subw_aggregator)
feminine = out.sequence_attributions[1].aggregate(aggregator=subw_aggregator)

# Take the diff of the scores of the two attributions, show it and return the HTML
html = masculine.show(aggregator=PairAggregator, paired_attr=feminine, return_html=True)
\end{lstlisting}
    \includegraphics[width=.5\textwidth]{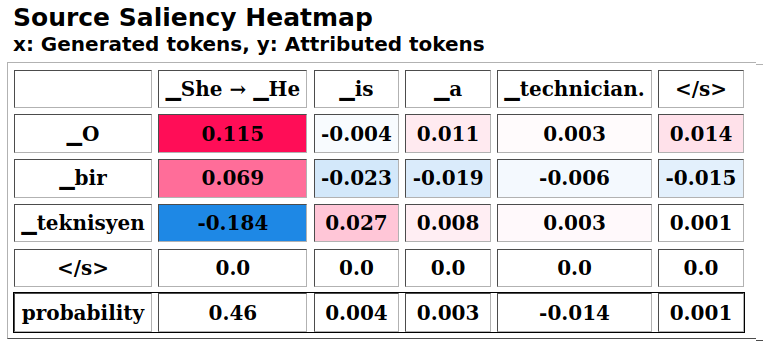}
    \caption{Comparing attributions for a synthetic Turkish-to-English translation example with underspecified source pronoun gender using a MarianMT Turkish-to-English translation model~\citep{tiedemann-2020-tatoeba}. Values in the visualized attribution matrix show a 46\% higher probability of producing the masculine pronoun in the translation and a relative decrease of 18.4\% in the importance of the Turkish occupation term compared to the feminine pronoun case.}
    \label{fig:ex-gender-pair-agg}
\end{figure*}

An example of gender translation pair using the synthetic template of Section~\ref{sec:gender-bias} is show in Figure~\ref{fig:ex-gender-pair-agg}, highlighting a large drop in probability when switching the gendered pronoun for highly gender-stereotypical professions, similar to Table~\ref{tab:turkish_gender_bias} results.

\section{Example of Quantized Contrastive Attribution of Factual Knowledge}
\label{app:ex-factual-int8}

\begin{figure*}[!t]
    \centering
    \begin{lstlisting}[language=Python]
import inseq
from datasets import load_dataset
from transformers import AutoModelForCausalLM, AutoTokenizer

# The model is loaded in 8-bit on available GPUs
model = AutoModelForCausalLM.from_pretrained("gpt2-xl", load_in_8bit=True, device_map="auto")
tokenizer = AutoTokenizer.from_pretrained("gpt2-xl")
# Counterfact datasets used by Meng et al. (2022)
data = load_dataset("NeelNanda/counterfact-tracing")["train"]

# GPT-2 XL is a Transformer model with 48 layers
for layer in range(48):
    attrib_model = inseq.load_model(
        model,
        "layer_gradient_x_activation",
        tokenizer="gpt2-xl",
        target_layer=model.transformer.h[layer].mlp,
    )
    for i, ex in data:
        # e.g. "The capital of Second Spanish Republic is"
        prompt = ex["relation"].format{ex["subject"]}
        # e.g. "The capital of Second Spanish Republic is Madrid"
        true_answer = prompt + ex["target_true"]
        # e.g. "The capital of Second Spanish Republic is Paris"
        false_answer = prompt + ex["target_false"] 
        contrast = attrib_model.encode(false_answer)
        # Contrastive attribution of true vs false answer
        out = attrib_model.attribute(
            prompt,
            true_answer,
            attributed_fn="contrast_prob_diff",
            contrast_ids=contrast.input_ids,
            contrast_attention_mask=contrast.attention_mask,
            step_scores=["contrast_prob_diff"],
            show_progress=False,
        )
    # Aggregation and plotting omitted for brevity
    ...
\end{lstlisting}
    \includegraphics[width=0.33\textwidth]{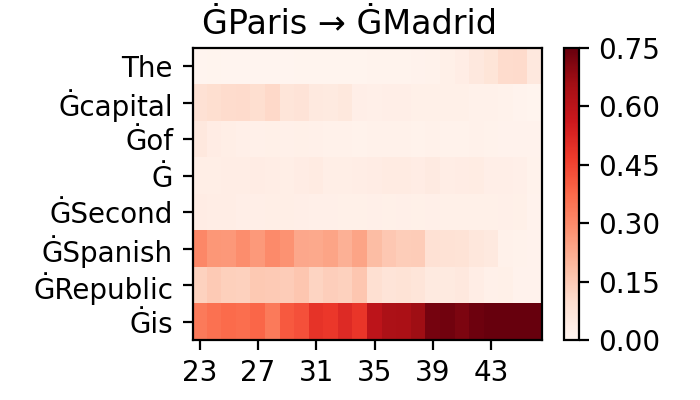}%
    \includegraphics[width=0.33\textwidth]{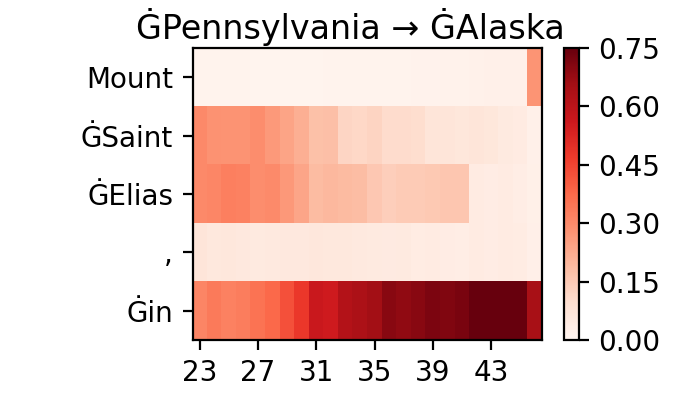}%
    \includegraphics[width=0.33\textwidth]{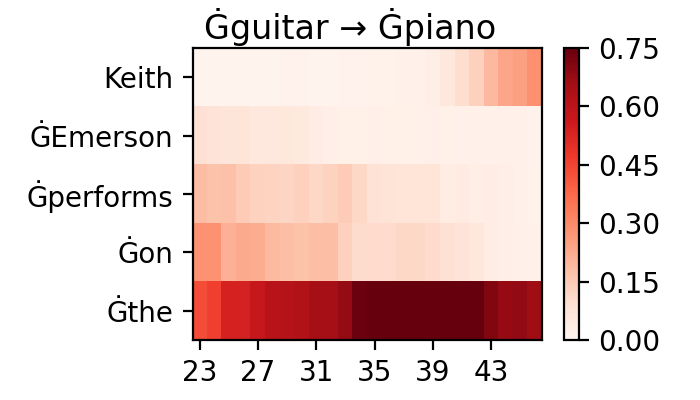}
    \includegraphics[width=0.33\textwidth]{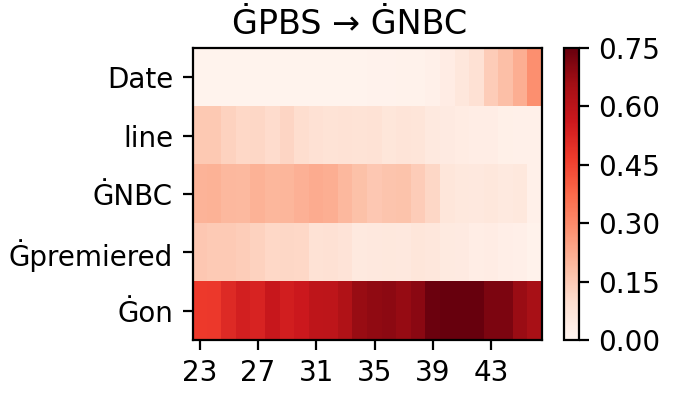}%
    \includegraphics[width=0.33\textwidth]{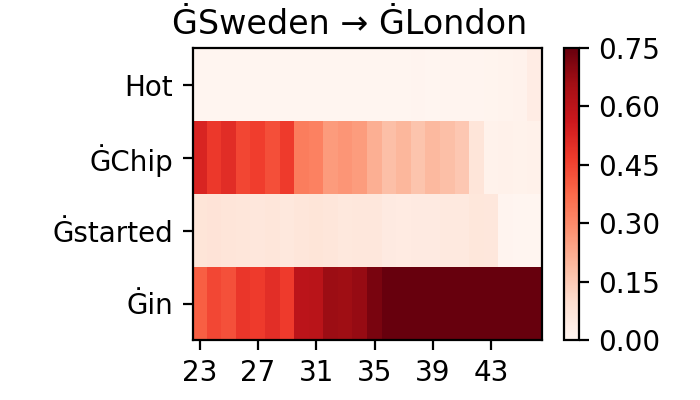}%
    \includegraphics[width=0.33\textwidth]{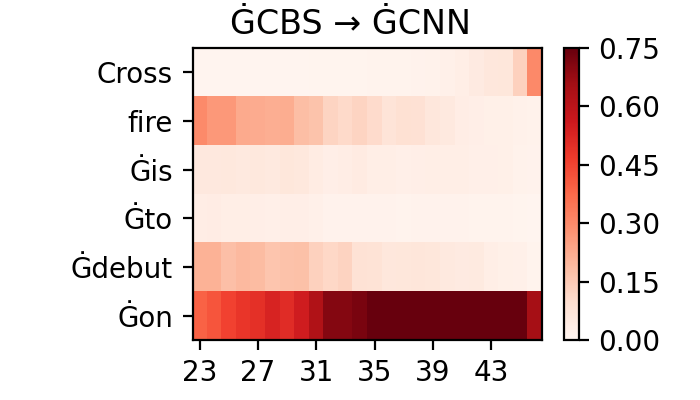}
    \caption{\textbf{Top:} Example code to contrastively attribute factual statements from the Counterfact Tracing dataset, using Layer Gradient $\times$ Activation to compute importance scores until intermediate layers of the GPT2-XL model. \textbf{Bottom:} Visualization of contrastive attribution scores on a subset of layers (23 to 48) for some selected dataset examples. Plot labels show the contrastive pairs of false $\rightarrow$ true answer used as attribution targets.}
    \label{fig:ex-factual-knowledge}
\end{figure*}

Figure~\ref{fig:ex-factual-knowledge} presents code used in Section~\ref{sec:rome-repro} case study, with visualized attribution scores for contrastive examples in the evaluated dataset.

\section{Gender Bias in Machine Translation}
\label{app:gender-bias-turkish}
Table \ref{tab:turkish-wordlist} shows the list of occupation terms used in the gender bias case study (Section \ref{sec:gender-bias}).
We correlate the ranking of occupations based on the selected attribution metrics and probabilities with U.S. labor statistics\footnote{\href{https://github.com/rudinger/winogender-schemas/blob/master/data/occupations-stats.tsv}{\texttt{https://github.com/rudinger/winogender-schemas}}} ({\small \texttt{bls\_pct\_female}} column).
Table~\ref{tab:m2m-gender-example} example was taken from the BUG dataset~\citep{levy-etal-2021-collecting-large}.

\begin{table}
\centering
\scriptsize
\begin{tabular}{ll|ll}
\toprule
Turkish       & English      & Turkish       & English      \\
\midrule
teknisyen     & technician   & memur         & officer      \\
muhasebeci    & accountant   & patolog       & pathologist  \\
süpervizör    & supervisor   & öğretmen      & teacher      \\
mühendis      & engineer     & avukat        & lawyer       \\
işçi          & worker       & planlamacı    & planner      \\
eğitimci      & educator     & yönetici      & practitioner \\
katip         & clerk        & tesisatçı     & plumber      \\
danışman      & consultant   & eğitmen       & instructor   \\
müfettiş      & inspector    & cerrah        & surgeon      \\
tamirci       & mechanic     & veteriner     & veterinarian \\
müdür         & manager      & kimyager      & chemist      \\
terapist      & therapist    & makinist      & machinist    \\
resepsiyonist & receptionist & mimar         & architect    \\
kütüphaneci   & librarian    & kuaför        & hairdresser  \\
ressam        & painter      & fırıncı       & baker        \\
eczacı        & pharmacist   & programlamacı & programmer   \\
kapıcı        & janitor      & itfaiyeci     & firefighter  \\
psikolog      & psychologist & bilim insanı  & scientist    \\
doktor        & physician    & sevk memuru   & dispatcher   \\
marangoz      & carpenter    & kasiyer       & cashier      \\
hemşire       & nurse        & komisyoncu    & broker       \\
araştırmacı   & investigator & şef           & chef         \\
barmen        & bartender    & doktor        & doctor       \\
uzman         & specialist   & sekreter      & secretary    \\
elektrikçi    & electrician  &               & \\
\bottomrule
\end{tabular}
    \caption{List of the 49 Turkish occupation terms and their English translations used in the gender bias case study (Section \ref{sec:gender-bias}).}
    \label{tab:turkish-wordlist}
\end{table}

\begin{table}[!ht]
\small
\centering
\begin{tabular}{p{0.2em}ll}
\toprule
& \textbf{Method} & \textbf{Source} \\
\midrule
\multirow{2}{*}{\textbf{G}} & Guided Integrated Gradients & \citeauthor{kapishnikov-2021-guided-ig} \\
& LRP & \citeauthor{bach-2015-lrp} \\
\midrule
\multirow{5}{*}{\textbf{I}} & Attention Rollout \& Flow & \citeauthor{abnar-zuidema-2020-quantifying} \\
& Attention $\times$ Vector Norm & \citeauthor{kobayashi-etal-2020-attention} \\
& Attention $\times$ Attn. Block Norm & \citeauthor{kobayashi-etal-2021-incorporating} \\
& GlobEnc & \citeauthor{modarressi-etal-2022-globenc} \\
& ALTI+ & \citeauthor{ferrando-etal-2022-towards} \\
& Attention $\times$ Trans. Block Norm & \citeauthor{kobayashi-etal-2023-feedforward} \\
& ALTI-Logit & \citeauthor{ferrando-etal-2023-explaining} \\
\midrule
\multirow{3}{*}{\textbf{P}} & Information Bottlenecks &  \citeauthor{jiang-etal-2020-inserting}\\
& Value Zeroing & \citeauthor{mohebbi-2023-value-zeroing} \\
& Input Reduction & \citeauthor{feng-etal-2018-pathologies} \\
& Activation Patching & \citeauthor{meng-2022-rome} \\
\bottomrule
\end{tabular}
\caption{Gradient-based (\textbf{G}), internals-based (\textbf{I}) and perturbation-based (\textbf{P}) attribution methods for which we plan to include support in future Inseq releases.}
\label{tab:planned-methods}
\end{table}

\section{Planned Developments and Next Steps}
\label{app:next-steps}

We plan to continuously expand the core functionality of the library by adding support for a wider range of attribution methods. Table~\ref{tab:planned-methods} shows a subset of methods we consider including in future releases. Besides new methods, we also intend to significantly improve result visualization using an interactive interface backed by Gradio Blocks \cite{abid-2019-gradio}, work on interoperability features together with ferret developers~\citep{attanasio-2022-ferret} to simplify the evaluation of sequence attributions, and include sequential instance attribution methods~\citep{lam-etal-2022-analyzing,jain-etal-2022-influence} for training data attribution.

\lstset{style=fullpagecode}

\lstset{style=fullpagecode}

\end{document}